\documentclass[Afour,sageh,times]{sagej}

\usepackage{moreverb,url}

\usepackage[colorlinks,bookmarksopen,bookmarksnumbered,citecolor=red,urlcolor=red]{hyperref}

\newcommand\BibTeX{{\rmfamily B\kern-.05em \textsc{i\kern-.025em b}\kern-.08em
T\kern-.1667em\lower.7ex\hbox{E}\kern-.125emX}}

\def\volumeyear{2016}

\usepackage{graphicx} 
\usepackage{placeins} 
\usepackage{textcomp} 
\usepackage[binary-units=true]{siunitx} 
\usepackage{multirow}

\DeclareSIUnit\px{px}

\begin{document}

\runninghead{Pire, Mujica, Civera and Kofman}

\title{The Rosario Dataset: Multisensor Data for Localization and Mapping in Agricultural Environments}

\author{Taih{\'u} Pire\affilnum{1}, Mart{\'i}n Mujica\affilnum{1}, Javier Civera\affilnum{2} and Ernesto Kofman\affilnum{1}}

\affiliation{\affilnum{1}CIFASIS, French Argentine International Center for Information and Systems Sciences (CONICET-UNR), Argentina\\
\affilnum{2}I3A, University of Zaragoza, Spain}

\corrauth{Taih{\'u} Pire, CIFASIS, French Argentine International Center for Information and Systems Sciences (CONICET-UNR), 27 de Febrero 210 bis, S2000EZP Rosario, Argentina}

\email{pire@cifasis-conicet.gov.ar}

\begin{abstract}
In this paper we present \emph{The Rosario Dataset}, a collection of sensor data for autonomous mobile robotics in agricultural scenes. The dataset is motivated by the lack of realistic sensor readings gathered by a mobile robot in such environments. It consists of 6 sequences recorded in soybean fields showing real and challenging cases: highly repetitive scenes, reflection and burned images caused by direct sunlight and rough terrain among others. The dataset was conceived in order to provide a benchmark and contribute to the agricultural SLAM/odometry and sensor fusion research. It contains synchronized readings of several sensors: wheel odometry, IMU, stereo camera and a GPS-RTK system. The dataset is publicly available in \url{http://www.cifasis-conicet.gov.ar/robot/}.
\end{abstract}


\maketitle

\section{Introduction}
Agriculture is one of the oldest and one of the most relevant industries for the human race. Its automation, with the goals of rising its productivity and releasing people from the most arduous tasks, has been a traditional line of research in the robotics community. 

The complete automation of agriculture, however, faces several challenges of great diversity. Among others, we can cite robot localization and environment mapping, weed/crop/fruit recognition, grasping and manipulation or navigation. The challenges might be different for each specific case, e.g., navigation will be easier in a crop field than through fruit trees. On the other hand, agricultural environments could be partially adapted to the robotic requirements if needed. In this regard, it might bear a resemblance to warehouse automation, and differs from service robots and automated cars, that face the additional challenge of adapting to human spaces.

Public datasets are an essential tool for the progress of a field. \cite{sturm2012benchmark,geiger2013vision} are two relevant examples related to visual SLAM and odometry. In this work, our aim is to contribute with the release of a public dataset for the tasks of localization and mapping in agricultural environments. The number of public datasets in the field of agricultural robotics, although it is increasing, is still insufficient.

Visual localization and mapping in agricultural environments presents a set of specific challenges; which are the motivation for the recording of this dataset. The most relevant ones are insufficient or repetitive texture --particularly challenging for loop closing--, small deviations from the rigid world assumption due to the wind, poor geometry --in many cases, just the flat ground plane--, irregular terrains --with the associated jumpy motion--, and a high variety of lighting conditions due to clouds passing or the direct view of the sun. We believe that our dataset is a significant contribution to benchmark existing algorithms for agricultural applications and develop new ones that are more suited to the particularities of agricultural scenes.

We organize the rest of the paper as follows: in Section \ref{sec:related} we present the most recent and relevant work related to datasets for agricultural robotics. In Section \ref{sec:robot}, we present the robot used to record the dataset along with its sensors and hardware configuration. In Section \ref{sec:dataset}, we describe the sensors calibration and the recording data methodology. Section \ref{sec:baselines} presents some experimental results, that illustrate the challenges and particularities of this dataset. In Section~\ref{sec:tools} we briefly describe some scripts used to record the dataset and to post-process the recorded data. In Section~\ref{sec:conclusions}, we summarize the conclusion and the future work.

\section{Related Work}
\label{sec:related}
The recent work \cite{chebrolu2017agricultural} is the public dataset most related to ours. It was recorded over 3 months on a sugar beet farm, and aims to advance research on crop/weed classification, localization and mapping. 

In the case of our dataset, we do not address long-term scene dynamics, and each recorded sequence corresponds to a different scene. As our aim is to evaluate localization and mapping capabilities, our data contains a wider array of scenes, with the aim of capturing a wider extent of challenging situations.

In the rest of this section we give an overview of several other related datasets. The survey is limited to the most recent and relevant works, with focus on visual data, and organized in two groups: First those works focusing in localization and mapping, and second those targeting agricultural applications.

\subsection{Datasets for localization and mapping}

There exists a high number of datasets for localization and mapping datasets. If we classify them by scene type, some of the most relevant are:

\begin{itemize}

\item \emph{Indoor scenes:} \cite{ruiz2017robot}, uses ground robots and with semantic annotations. \cite{burri2016euroc} is recorded from quadrotors and with geometric ground truth for the trajectory and the scene. \cite{sturm2012benchmark} is a classic one recorded with a RGB-D camera.

\item \emph{Outdoor urban scenes:} Some are taken by sensorized cars \cite{gpandey-2011a,blanco2014malaga}, others by small mobile robots \cite{smith2009new,carlevaris2016university} and others by quadrotors \cite{majdik2017zurich}. The recent \cite{maddern20171} is focused on life-long mapping from car vehicles.

\item \emph{Outdoor rural scenes:} \cite{miller2018visual} releases visual-inertial sequences taken from a canoe navigating along a river. \cite{griffith2017symphony} is  composed of several surveys over several years of a lake from an autonomous surface vehicle. \cite{leung2017chilean} presents data of an underground mine.

\item \emph{Simulated planetary scenes:} From all the existing ones, we can name for example \cite{furgale2012devon,tong2013canadian}

\end{itemize}

\subsection{Datasets for agricultural applications}

The number of datasets that are specifically related to agricultural robotics is less than those devoted to odometry and SLAM. However, recently, there is an emergence of several relevant ones, due to the growing importance of this application. In the next paragraph we mention several recent ones that have very different aims, reflecting the wide array of challenges in agricultural robotics. 

\cite{haug2014crop} presents a dataset containing real images for weed/crop classification. Due to the difficulty of scaling the dataset size with manual annotation, \cite{dpgp_IROS2017} addresses the same problem creating a synthetic dataset. \cite{sa2018weednet} addresses weed/crop classification from  multispectral images recorded by a MAV, using a deep neural network and releasing the data used. \cite{pezzementi2017comparing} targets person detection in off-road and agricultural scenes. \cite{fentanes20183d} contains soil compaction data, with the aim of advancing on autonomous soil compaction mapping by robots. \cite{dias2018multispecies} addresses the automated perception of bloom intensity --the number of flowers in an orchard--, which should guide operations like pruning and thinning for the desired fruit features. It releases an annotated dataset with pixelwise flower labels in HR images. In \cite{alencastre2018robotics} a dataset is collected for classifying the damage that automated harvesting caused in sugarcane billets. 




\section{The Weed Removing Robot}
\label{sec:robot}
In this section we describe the robot that we used to record the dataset, along with its sensors.

\subsection{The Robot}
The robot consists of a mobile platform with four wheels (see a picture in Figure~\ref{fig:robot_sensors}). It has been designed to work autonomously in large areas; and hence its power source are four batteries that are charged by photovoltaic cells in the top of the vehicle.

The robot has been designed to automate the weed removing tasks in large crop fields. Our aim is that the weeds and the crops are classified using visual data, and a tool (currently being developed) removes the weed without damaging the crops. The robot should navigate through the field autonomously and should keep track of the state of each land piece; hence it needs accurate localization and mapping capabilities in this environment.

The robot motion is controlled by four brush-less motors (one per wheel) and their drivers. For the front wheels direction, a stepper motor has been built with the appropriate reduction and encoder.

\subsection{The Sensors}
\begin{figure}[!htb]
\centering
\includegraphics[width=\columnwidth]{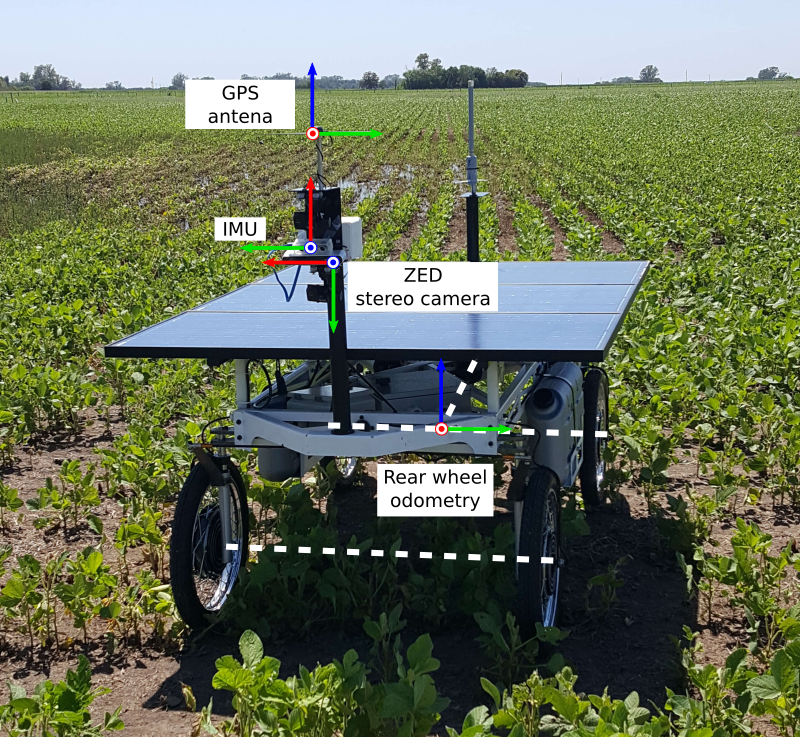}
\caption{The weed removing robot and its sensors. The right camera coordinate frame is not shown for clarity.}
\label{fig:robot_sensors}
\end{figure}

The picture in Figure~\ref{fig:robot_sensors} shows the sensors that are mounted in the robot and their respective local frames. The main technical details of the sensors are as follows:

\begin{itemize}

\item{\bf Stereo Camera.} We used the \emph{ZED stereo camera}\footnote{\url{https://www.stereolabs.com/zed/}}. The camera baseline is \SI{12}{\centi\meter}. We recorded synchronized left and right images at a resolution of $672\times\SI{376}{\px}$, and at a frame rate of \SI{15}{\hertz}. 

\item{\bf Motors encoders} We used three Hall effect sensors coupled with each wheel to measure rotational angle increments. From this data, using a kinematic model of our robot, we extracted its linear and angular motion. 

\item{\bf GPS-RTK}. We used two \emph{GPS-RTK Reach}\footnote{\url{https://emlid.com/reach/}} modules, one mounted on the robot and another one on the base station. The GPS-RTK frequency is \SI{5}{\hertz}. Its accuracy was characterized in our previous work \cite{pistarelli2017gps}. The base station consists of a Reach module connected to a Tallysman TW4721 antenna with IP67 protection. It is mounted on a $322\times\SI{247}{\mm}$ ground plane that far exceeds the $100\times\SI{100}{\mm}$ suggested by the manufacturer of the GPS-RTK, giving it superior rejection of bouncing signals from nearby structures (\emph{multipath signals}). The connection between the two GPS-RTK modules was made through a WiFi network using two routers. The first one, a MikroTik Metal G-52SHPacn, was placed in a fixed base station, while the second one, a MikroTik Groove GA-52HPacn, was placed on the robot. The routers were chosen due to their high transmission power and receiver sensitivity. The main difference between them is that the one placed on the robot, has a lower power consumption. 

In order to energize both systems, a module powered by four rechargeable lithium cells with a total capacity of \SI{10.400}{\milli\ampere\hour} was chosen to integrate the \emph{switching} regulated charge and output system of \SI{5}{\volt} voltage in the same container.

\item{\bf Inertial Measurement Unit.} The IMU that we used is the LSM6DS0, that is built in the \emph{TARA stereo-inertial sensor}\footnote{\url{https://www.e-consystems.com/}}. The IMU rate was set to \SI{140}{\hertz}.
\end{itemize}

Although our sensor equipment included the TARA stereo camera, its auto-exposure setting was not appropriate for our outdoor scenes and produced burned images. We then had to discard the images, but we kept the IMU data.

The ZED stereo camera, IMU and motors encoders were connected by USB 3.0 to the robot's computer. The GPS-RTK information was read through WiFi network.

\subsection{The Computer}
All the sensor data we recorded was timestamped and stored in an onboard robot computer. We used a MINI-PC {Intel\textregistered} NUC Kit NUC6CAYH\footnote{\url{www.intel.com}} (Intel Celeron J3455 CPU, quad-core \SI{2.3}{\giga\hertz} and \SI{8}{\giga\byte} DDR3 RAM Memory) with Ubuntu 16.04. The supplied voltage (\SI{12}{\volt})  came from the robot batteries.
In order to record data as soon as it arrives to the Operative System and avoid disk-writing delays, we used a Solid State Disk (specifically a \SI{240}{\giga\byte} Western Digital SSD WDS 240G1G0A) as storage unit. 

\section{The Dataset}
\label{sec:dataset}
In this section we detail the calibration of the robot sensors and the format of the recorded data.

\subsection{Calibration}
The extrinsic and the intrinsic calibrations of all sensors (for each sequence) are stored in the files \texttt{calibration.txt} (Table~\ref{tab:extrinsic_parameters} shows the extrinsic parameters). The calibration file includes the camera and IMU intrinsic parameters; and the transformations between all the sensors.
\begin{table*}[!htb]
  \centering
  \begin{tabular}{ccccccccc}
    \toprule
    Frame ID & Child Frame ID & $x \left[\si{\meter}\right]$ & $y \left[\si{\meter}\right]$ & $z \left[\si{\meter}\right]$ & $q_{x}$ & $q_{y}$ & $q_{z}$ & $q_{w}$  \\
    \midrule
    rear\_wheel\_odometry & base\_link & 0 & 0 & 0 & 0 & 0 & 0 & 1\\
    \hline
    base\_link & gps & 1.8 & -0.030 & 1.593 & - & - & - & -\\
    \hline
    imu & left\_camera & -0.031 & -0.077 & 0.026 & 0.058 & 0.019 & 0.703 & 0.708\\
    \hline
    imu & right\_camera &-0.030 & 0.042 & 0.033 & 0.064 & 0.012 & 0.703 & 0.708\\
    \bottomrule
  \end{tabular}
  \caption{Rigid transformations between the different coordinate frames involved in the system. The translation is given by $x$, $y$ and $z$; and the rotation by the quaternion $q$. The transformations are defined to convert points from the Child Frame ID to the Frame ID. The rigid transformation between the base\_link and the imu coordinate frames changes for each sequence and therefore is not shown in the table.}
  \label{tab:extrinsic_parameters}
\end{table*}
\subsubsection{Intrinsic parameters}
For each camera of the ZED stereo we used a standard pinhole model with radial-tangential distortion. We calibrated the intrinsics of each camera with Kalibr \cite{maye2013self}.

We used the \emph{Allan variance} method~\cite{allan1966statistics} for estimating the IMU noise model. The noise model is given by accelerometer noise density ($\sigma_{g}$), accelerometer random walk bias ($\sigma_{bg}$), gyroscope noise density ($\sigma_{a}$), gyroscope random walk bias ($\sigma_{ba}$) and the sampling rate $\left( \frac{1}{\Delta t} \right)$. The specific values for the IMU noise model are in Table~\ref{table:imu_calibration}.
\begin{table}[!htb]
  \centering
  \begin{tabular}{lcc}
  Parameter & Value & Units\\
  \midrule
  $\dfrac{1}{\Delta t}$ & $142.0$ & $\si{\hertz}$\\
  $\sigma_{g}$ & $8.2739$
  & $\si{\frac{\radian}{\second} \frac{1}{\sqrt{\hertz}}}$ \\
  $\sigma_{bg}$ & $8.7367$
  & $\si{\frac{\meter}{\second^{2}} \frac{1}{\sqrt{\hertz}}}$ \\
  $\sigma_{a}$ & $0.0017$
  & $\si{\frac{\radian}{\second^{2}} \frac{1}{\sqrt{\hertz}}}$ \\
  $\sigma_{ba}$ & $0.0057$
  & $\si{\frac{\meter}{\second^{3}} \frac{1}{\sqrt{\hertz}}}$ \\
  \bottomrule
\end{tabular}
  \caption{IMU calibration parameters.} \label{table:imu_calibration}
\end{table}
\subsubsection{Extrinsic parameters}
We chose the local frame of the rear\_wheel\_odometry as our robot base\_link, and referenced the extrinsics of all the other sensors to such frame. We used Kalibr \cite{furgale2012continuous,furgale2013unified} to calibrate the stereo extrinsics (the relative pose between the left and right cameras) and the relative transformation between the cameras and the IMU. 

We calibrated the rigid transformation between the odometry coordinate frame (rear\_wheel\_odometry) and the left camera frame (left\_camera) as follows. 
First, we estimated the motion of the robot referred to the left camera frame in small straight segments of our data.
We used the stereo SLAM system S-PTAM \cite{pire2017sptam} for that. Then, we averaged the normalized estimated positions to estimate the local motion vector. We calculated the rotation matrix between such motion vector and the forward axis of the camera $\left[0, 0, 1\right]$. We denote this rotation as $\mathbf{R}(\theta)$, being $\theta$ the angle between the motion and the camera z-axis. The $\mathbf{R}(\theta)$ rotation, composed with the \SI{90}{\degree} rotations required to align the axis of both frames, form the rotation between the odometry and the left camera coordinate frames, as detailed in equation \ref{eq:odometry_camera_rotation}.

\begin{equation}
    \mathbf{R}^{\mathrm{r}\mathrm{c}} = \mathbf{R}_{z}({\SI{90}{\degree}}) \mathbf{R}_{x}({\SI{90}{\degree}}) \mathbf{R}(\theta),
    \label{eq:odometry_camera_rotation}
\end{equation}

where $\mathbf{R}^{\mathrm{r}\mathrm{c}}$ stands for the rotation matrix between the left camera frame and the odometry frame. The rotation matrices $\mathbf{R}_{z}({\SI{90}{\degree}})$ and $\mathbf{R}_{x}({\SI{90}{\degree}})$ are the \SI{90}{\degree} rotations around the $z$ and $x$ axis respectively. 
We estimated the translation part of the rigid transformation between both sensors by directly measuring them on the robot.

To keep the left and right camera as child frames of the IMU frame, we used the transformation between the odometry frame and the left camera frame ($ \mathbf{T}^{\mathrm{r}\mathrm{c}}$), along with the transformation between the the left camera and the IMU ($\mathbf{T}^{\mathrm{c}\mathrm{i}}$) to calculate the relative pose between the odometry frame and the IMU frame $\mathbf{T}^{\mathrm{r}\mathrm{i}}$, which is the one used in the transformations tree.

\begin{equation}
    \mathbf{T}^{\mathrm{r}\mathrm{i}} = \mathbf{T}^{\mathrm{r}\mathrm{c}} \mathbf{T}^{\mathrm{c}\mathrm{i}}.
    \label{eq:odometry_imu_transformation}
\end{equation}

The relative transformation between the rest of the sensors (GPS-RTK and odometry) was calibrated using AprilTags \cite{olson2011apriltag}. We attached the tags to the sensors and estimated their relative transformations from multiple views taken by an external camera. In particular, we used the ar\_track\_alvar\footnote{\url{http://wiki.ros.org/ar_track_alvar}} ROS package.

\subsection{Data synchronization}
As we are working with end-user sensors (ZED stereo camera, TARA visual-inertial sensor and GPS-RTK Reach modules), all data was synchronized by software at the level of user applications in the Operative System. The data was straightforwardly recorded on a solid state drive in the robot on-board computer. We use a precision of milliseconds for measurement timestamps labeled.

\subsection{Data Collection and Summary}
The data was collected in two separate days in the agriculture fields used by the Faculty of Agricultural Science at the National University of Rosario, Argentina. We recorded $6$ sequences, with a total trajectory length around $2.3$ kilometers and a total time around $30$ minutes.
\begin{table*}[!htb]
  \centering
  \small
  \begin{tabular}{llllll}
  \toprule
  Sequence \# & Difficulty & Length (\si{\meter}) & Duration (\si{\minute}) & Sequence ID (date\_time) & Summary\\
  \midrule
  1 & easy & 615.15 & 9.3 & 2017-12-26\_12:25:45 & $3 \times 180^{\circ}$ turn\\
    & & & & & Occasional backwards motion\\
    & & & & & People (occasional)\\
    & & & & & Partial occlusions\\
    & & & & & Easily visible furrows\\
    & & & & & Green crops\\
  \midrule
  2 & easy & 320.16 & 4.4 & 2017-12-29\_11:13:55 & $1 \times 180^{\circ}$ turn\\
  & & & & & Dried crops\\
  & & & & & People (occasional)\\
  & & & & & Hardly visible furrows\\
  \midrule
  3 & medium & 169.45 & 3.3 & 2017-12-29\_11:23:00-part1 & $1 \times 180^{\circ}$ turn\\
  & & & & & Occasional backwards motion\\
  & & & & & Dried and green crops\\
  & & & & & Easily visible furrows\\
  \midrule
  4 & medium & 152.32 & 2.7 & 2017-12-29\_11:23:00-part2 & No turns\\
  & & & & & Easily visible furrows\\
  & & & & & Green crops\\
  \midrule
  5 & difficult & 330.43 & 5.2 & 2017-12-29\_11:47:35 & $1 \times 180^{\circ}$ turn\\
  & & & & & Occasional backwards motion\\
  & & & & & Varied furrow visibility\\
  & & & & & People (occasional)\\
  \midrule
  6 & difficult & 709.42 & 9.8 & 2017-12-29\_12:00:07 & $2 \times 180^{\circ}$ turn\\
  & & & & & People (occasional)\\
  & & & & & Road crossing\\
  \bottomrule
\end{tabular}
  \caption{Sequences description.} \label{table:sequence_description}
\end{table*}
\begin{figure}[!htb]
\centering
\includegraphics[width=\linewidth]{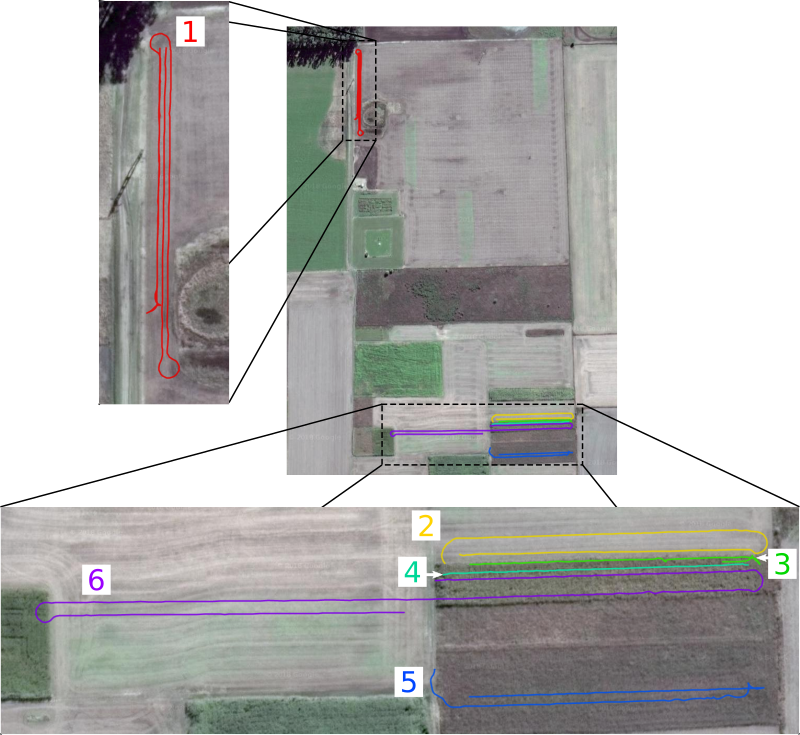}
\caption{GPS-RTK trajectories for the $6$ sequences of the dataset.}
\label{fig:trajectories}
\end{figure}

Figure~\ref{fig:trajectories} shows the GPS-RTK trajectories of the $6$ experiments. We commanded the robot to navigate along the furrows, with $180^\circ$ turns at their ends. Such trajectories are the less damaging for the crops, so we assume robots similar to ours will follow similar ones in agricultural applications. Due to the non-holonomic constraints of our platform and distance between furrows, $180^\circ$ turns require maneuvering and short backwards motions.

Table~\ref{table:sequence_description} contains more technical details for each of the sequences and a qualitative grade (from easy to difficult) and summary. The grade is based on our visual inspection and the results offered by visual SLAM baselines (see section~\ref{sec:baselines}). The data was recorded aiming to show a high variety of conditions in the fields: From green to dried crops, and from low to high vegetation density (that makes the furrows more or less visible). Such variations are reported in the table.

In addition to the particularities of each sequence, the data presents the challenges associated with agricultural applications mentioned in the introduction. The feature density is irregular. Visual tracking is difficult, due to texture similarities and non-rigid motions. The latest are mainly caused by light wind, and also by people that occasionally enter the field of view of the cameras. The robot motion is bumpy due to the uneven terrain, which makes tracking harder. The rolling shutter of the \emph{ZED stereo camera} adds an extra complexity, but we believe that such cameras are the most reasonable option for massive robot deployment due to its low cost.

\begin{figure*}[!htb]
\centering
\includegraphics[width=\linewidth]{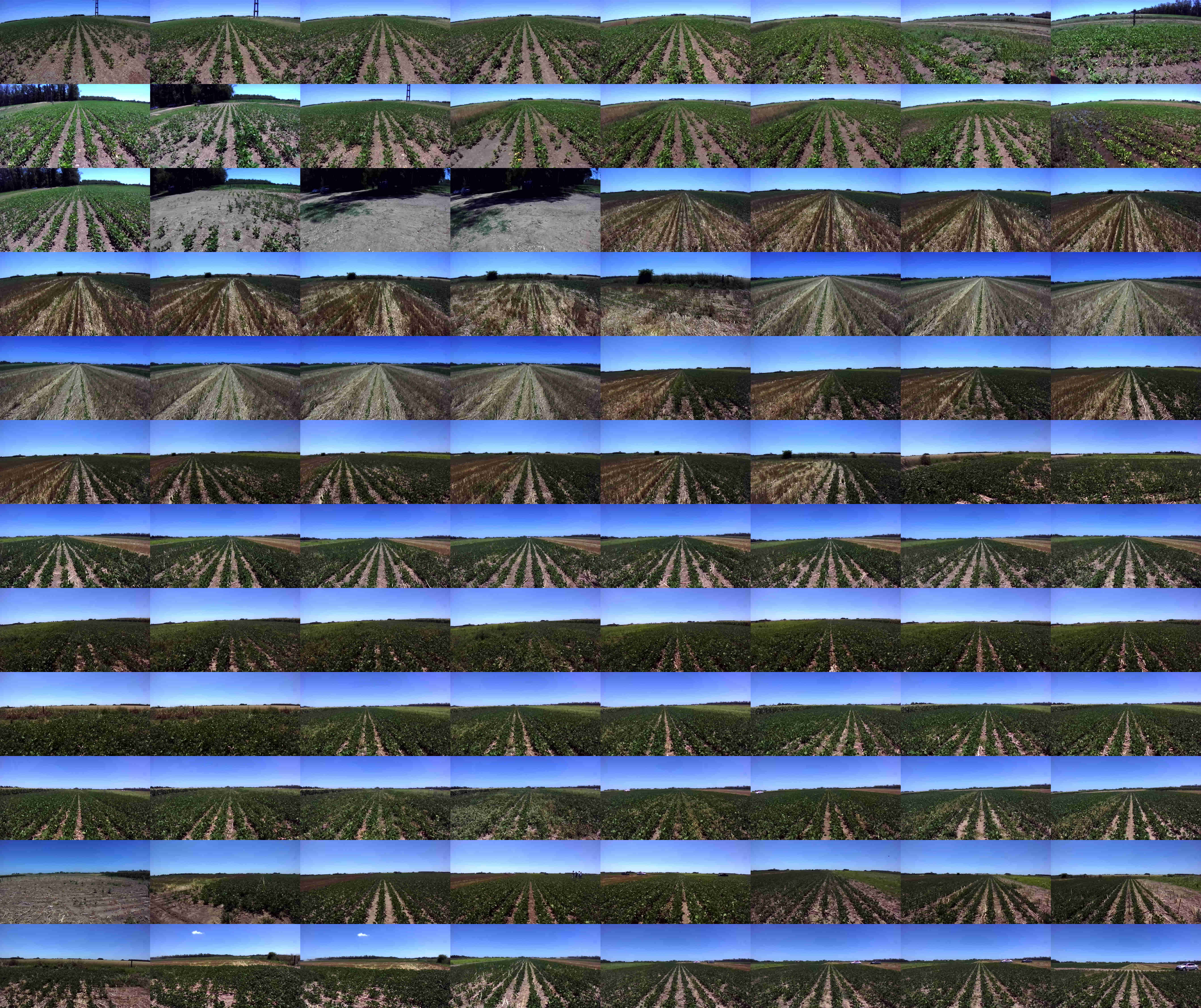}
\caption{Sample images of all the sequences of the dataset.}
\label{fig:dataset_samples}
\end{figure*}

Figure~\ref{fig:dataset_samples} shows several sample images from all the sequences of the dataset. Notice the mentioned variability in the crop and field conditions, the low texture and the repetitive patterns.

\subsection{Data Formats}

\begin{figure}[!htb]
\centering
\includegraphics[width=0.6\linewidth]{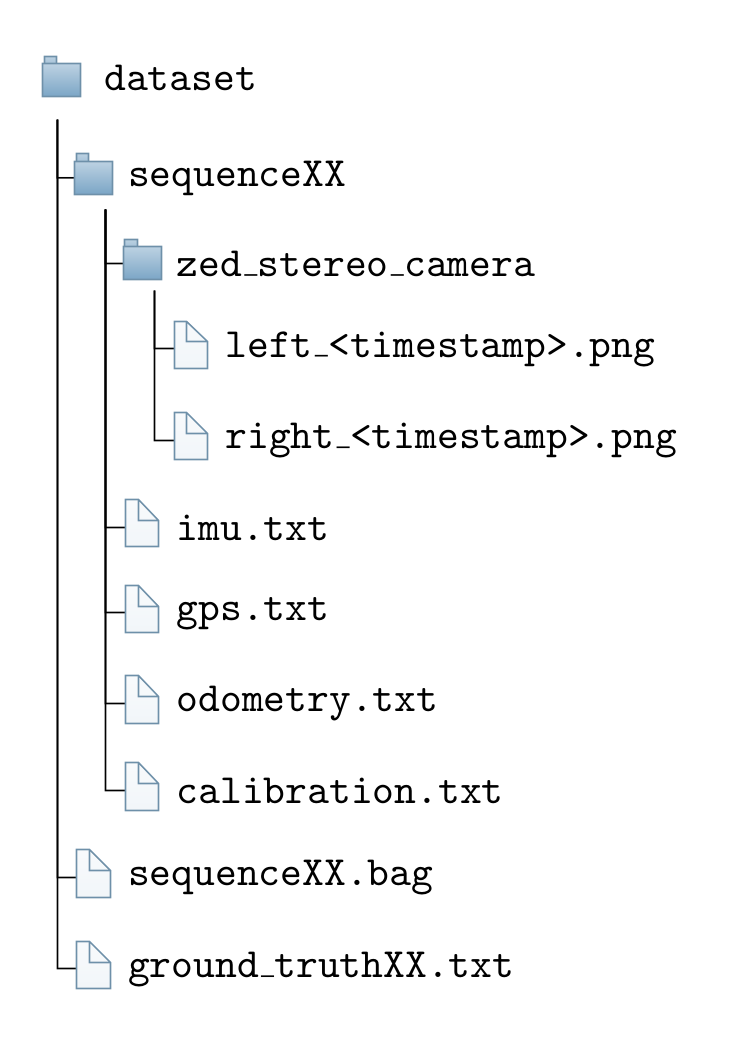}
\caption{Dataset structure. The suffix XX references to all the sequences numbers (01..06).}
\label{fig:dataset_structure}
\end{figure}

Figure~\ref{fig:dataset_structure} shows the dataset folder structure. We included the raw data and also the processed rosbags containing standard ROS messages, in order to facilitate its use. The data is, specifically, stored as follows.

\subsubsection{Raw Data}
\begin{itemize}
\item{\bf ZED stereo camera:} There is a folder containing both left and right images in \texttt{.png} format. The image size is $672\times\SI{376}{\px}$ and the naming convention is \texttt{\textless camera\textgreater \_\textless timestamp\textgreater.png}
\item{\bf IMU}: The file contains the measurements of the angular velocity and linear acceleration along the three axis as \texttt{\textless timestamp\textgreater} \texttt{\textless gyro[x,y,z]\textgreater} \texttt{\textless acc[x,y,z]\textgreater}
\item{\bf GPS-RTK:} The data follows the NMEA standards, giving the traditional latitude-longitude information, but also ground speed and satellites status. Each NMEA sentence has its own timestamp. 
\item{\bf Odometry:} We record the information from the wheel motors at a frequency of \SI{10}{\hertz}, along with the current timestamp. This information consists of the linear velocity of each motor and the current angle of the stepper motor that drives the direction. 
The measurements are contained in a file where each line is structured as
\texttt{\textless timestamp\textgreater} \texttt{\textless vel\_left\textgreater} \texttt{\textless vel\_right\textgreater} \texttt{\textless angle\textgreater} \texttt{\textless direction\textgreater}.
\end{itemize}

\subsection{Rosbags}
In addition to the raw data, we provide a rosbag for each sequence, the data being adapted to fit into ROS standard messages. This allows the use of the dataset with ROS-based software with minimum overload. The type of messages included in the \texttt{.bag} files are:

\begin{itemize}
\item {\bf sensor\_msgs/Image.msg}. Left and right images from the ZED stereo camera. 
\item {\bf sensor\_msgs/CameraInfo.msg}. Intrinsic and extrinsic parameters of both cameras. The right camera pose is referred to the left camera coordinate system.
\item {\bf sensor\_msgs/Imu.msg}. Raw IMU measurements. 
\item {\bf sensor\_msgs/NavSatFix.msg}. We publish the ``GGA'' part of the NMEA sentences provided by the GPS-RTK. The GGA sentence includes positioning and its estimated accuracy.
\item {\bf nav\_msgs/Odometry.msg}. Linear and angular velocity derived from the wheel encoders and the robot kinematic model. We also publish the integrated pose, resulting from the integration of the velocities.
\item{\bf tf/tfMessage.msg}. Extrinsic transformations between coordinate systems (see Table~\ref{tab:extrinsic_parameters}). All the extrinsic transformations between sensors are expressed as a rotation quaternion and a 3D translation vector. Since the rear\_wheel\_odometry and the base\_link coordinate systems are coincident, we publish the odometry messages on the base\_link system and remove the rear\_wheel\_odometry frame from the {\bf tf} message for clarity.
\end{itemize}

\subsection{Wheel odometry}


We generated the robot wheel odometry using the Ackerman model \cite{weinstein2010pose}. The wheelbase of our robot is $\SI{1.6}{\meter}$, the steering angle $\delta \in \left[-19,19\right]$ degrees and the wheel diameter $\SI{0.57}{\meter}$. Notice that the dataset includes the post-processed odometry and the raw one, directly read from the sensors, in case other kinematic model is preferred.

\subsection{Ground Truth}
We provide a positional GPS-RTK ground truth in order to assess the VO and SLAM accuracies. Since the IMU does not have a magnetometer, no global orientation is provided.

As having the ground-truth data in the the robot frame (base\_link) is necessary for comparing the trajectories, we computed the rotation between the robot trajectory and the GPS-RTK positions in small data subsets (less than \SI{10}{\meter}) of each sequence, where the robot is approximately moving in a straight line. We obtained the trajectory performed by the robot using the visual SLAM system S-PTAM \cite{pire2015sptam,pire2017sptam} which provides a highly accurate pose in highly textured environments. Observe that S-PTAM has been run offline in order to guarantee the best performance.

The rotation transformation between both trajectories is computed using the Horn method provided in \cite{sturm2012benchmark} and applied to the original GPS data to obtain the ground-truth presented in the dataset.





\section{Baselines}
\label{sec:baselines}
We run two state-of-the-art baselines for stereo SLAM, ORB-SLAM2 \cite{mur2017orb} and S-PTAM \cite{pire2017sptam}, in order to illustrate the characteristics and challenges of our dataset. Both systems were run with their default configuration. Table \ref{tab:orbslamsptam} shows the absolute trajectory error (ATE, as defined in \cite{sturm2012benchmark}) and, in brackets, the ratio of such error over the trajectory length. For comparison, we also show the same metrics for both system in three sequences of KITTI dataset \cite{geiger2013vision}, comparable in length to ours.

\begin{table}[!htb]
  \centering
  \begin{tabular}{ccc}
    \toprule
    Sequence & ORB-SLAM2 & S-PTAM \\ 
    \hline
    Rosario 01 & 1.41 (0.23\%) & 3.85 (0.63\%) \\
    Rosario 02 & 2.24 (0.70\%) & 1.80 (0.56\%) \\
    Rosario 03 & 3.50 (2.06\%) & 2.37 (1.40\%) \\
    Rosario 04 & 2.21 (1.45\%) & 1.49 (0.98\%) \\
    Rosario 05 & 2.23 (0.68\%) & X \\
    Rosario 06 & 5.19 (0.73\%) & X \\
    \hline
    KITTI 03 & 0.60 (0.11\%) & 1.66 (0.30\%) \\
    KITTI 05 & 0.80 (0.04\%) & 2.85 (0.13\%) \\
    KITTI 06 & 0.80 (0.06\%) & 2.99 (0.24\%) \\
    \bottomrule
  \end{tabular}
  \caption{Absolute trajectory error (ATE) [m] (ratio ATE over trajectory length, in \%), for ORB-SLAM2 and S-PTAM in the sequences of the Rosario dataset and a selection of KITTI (X stands for tracking failure).}
  \label{tab:orbslamsptam}
\end{table}

Notice how the error ratios for the Rosario dataset are significantly higher than the ones using the KITTI sequences. The challenges mentioned in the introduction (insufficient and repetitive texture, non-rigid motion of the plants, lighting changes and jumpy motion) cause the rapid loss of the feature tracks. As a consequence, among others, of the small length of the feature tracks, the drift grows quickly and failures are most likely.

\begin{figure}[!htb]
\centering
\includegraphics[width=.85\columnwidth]{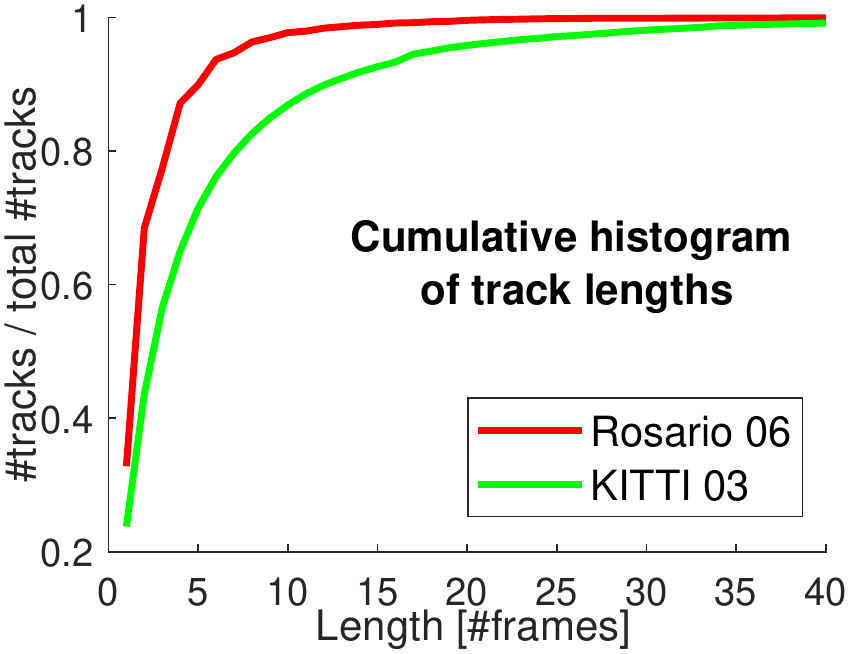}
\caption{Cumulative histogram of the length of S-PTAM feature tracks, for a representative sequence of the Rosario and KITTI datasets.}
\label{fig:cumhisttracks}
\end{figure}

Figure \ref{fig:cumhisttracks} shows a cumulative histogram of the length of features tracked by S-PTAM in two representative sequences, $06$ from the Rosario dataset and $03$ from KITTI. Notice the higher amount of small-length tracks in the Rosario sequence, illustrating the challenges in having high-quality and long feature tracks in agricultural scenes.

\begin{figure}[!htb]
\centering
\includegraphics[width=\columnwidth]{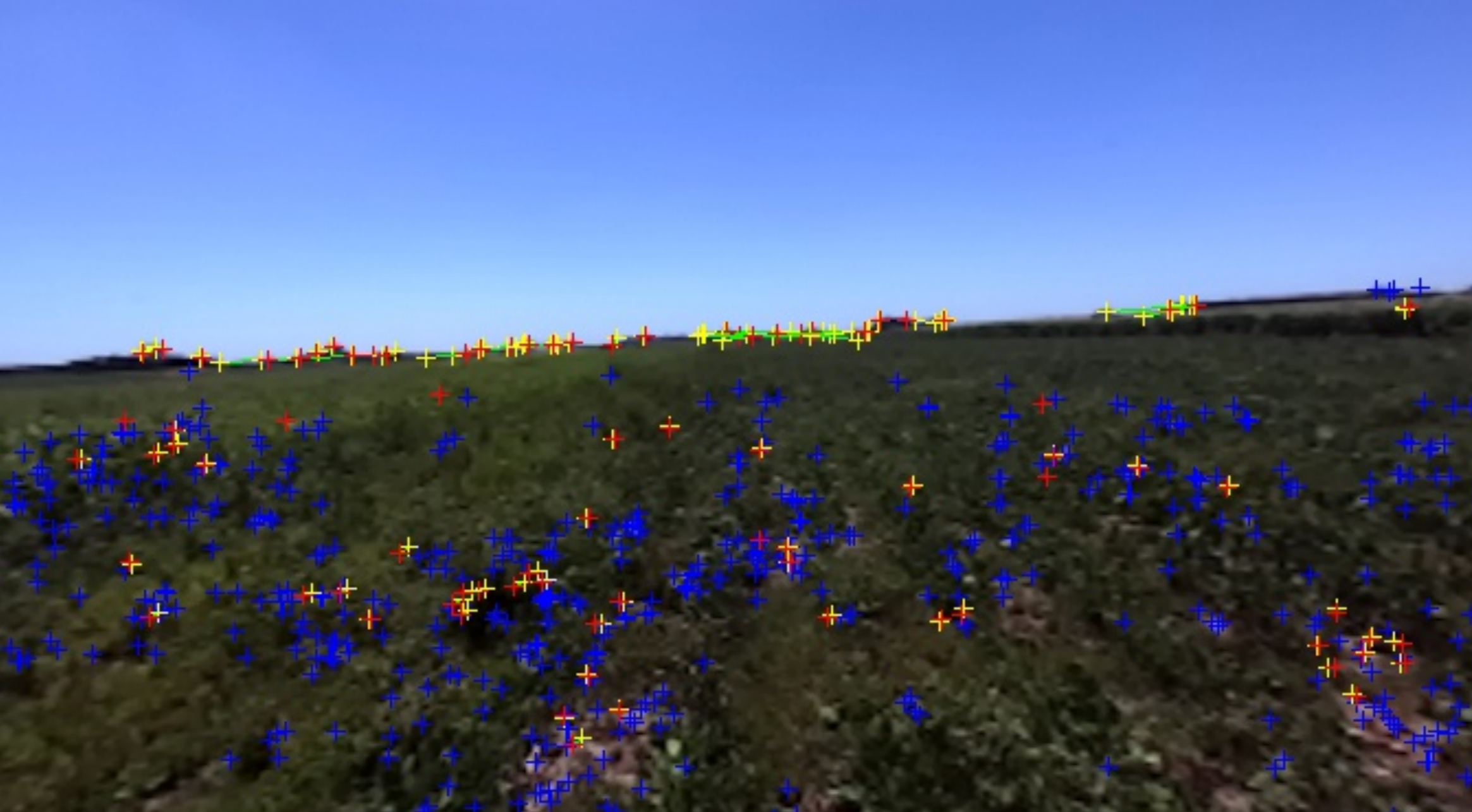}
\caption{Feature tracks example. In yellow, reprojection of map points tracked in this frame. In red, point correspondences. In blue, map points that cannot be tracked. Notice the high number of these latest ones.}
\label{fig:imagetracks}
\end{figure}

Finally, Figure \ref{fig:imagetracks} shows a representative frame of our dataset where we can see the tracked features. Notice, first, in blue, the high number of map points that cannot be matched in this particular frame. Observe also that the number of features tracked (matches in red, map point projections in yellow) is moderate. As mentioned, this small number of tracks and its small duration causes drift.

\section{Development Tools}
\label{sec:tools}
In order to access the robot sensors and collect the raw data, we developed several tools that capture the frames from the ZED stereo camera and the data from the IMU, the GPS-RTK and the motor encoders. In the camera recording software we decoupled the image recording and writing processes, to avoid losing frames. We implemented an application using a producer-consumer multi-thread architecture. This is, one thread is in charge of reading the images captured by the camera and pushing them in a FIFO queue. A second thread pulls the images from the queue, in the order they were stored, and saves them on the disk.

We developed a set of python scripts to generate the ROS messages from the raw data and to parse the calibration parameters from one format to another. We summarize here two of the script most relevant for the processing of the data:

\begin{itemize}
	\item{\it create\_bagfile.py} generates a rosbag from the raw data recorded by all of our sensors. It also consider the intrinsics and extrinsics calibration parameters to generate the CameraInfo messages for each camera.
    \item{\it imu\_convertion.py} processes the IMU data in order to have the acceleration expressed in $\si{\frac{\meter}{\second^{2}}}$ and the angular velocity in $\si{\frac{\radian}{\second}}$. This script also removes the offsets estimated by the Kalibr tool calibration.
\end{itemize}

We use the \emph{Allan Tools} software\footnote{\url{https://github.com/GAVLab/allan_variance}} to obtain the IMU noise model through the \emph{Allan variance}.

All the tools described are provided, along with the dataset. The aim is to allow and facilitate the manipulation of the raw data to replicate our results, to obtain new ones and to help in the recording of new datasets with the sensors we used.
 

\section{Conclusions and Future Work}
\label{sec:conclusions}
The aim of this work is the public release of a dataset, as a tool for other researchers to evaluate and improve their algorithms. We target the SLAM and odometry communities working with visual sensors (the dataset contains calibrated stereo data) and with fusion of odometric, inertial and visual information. 

The sequences were recorded in large agricultural environments, a non-traditional scenario for localization and mapping where few datasets exist. The monotony of the surroundings of a robot and the lack of texture are challenges for its visual positioning, that are present in our dataset. We believe that our dataset will contribute to the development of methodologies and algorithms suitable for such an important area of work as agriculture.


\begin{acks}
This work is part of the \textit{Development of a weed remotion mobile robot} project at CIFASIS (CONICET-UNR). It was also partially supported by the Spanish government (project DPI2015-67275) and the Arag\'on regional government (Grupo DGA-T45\_17R/FSE).
\end{acks}

\bibliographystyle{SageH} 
\bibliography{biblio}

\end{document}